\documentclass[letterpaper, 10 pt, journal, twoside]{IEEEtran}
\ifCLASSINFOpdf
\else
\fi

\usepackage{amssymb,amsmath}
\usepackage{cite} 
\usepackage{booktabs}
\usepackage{array}
\usepackage{diagbox}
\usepackage{graphicx}
\usepackage[font=footnotesize,labelformat=simple]{subcaption}
\usepackage[font=footnotesize]{caption}

\newcommand{\specialcellbold}[2][c]{%
  \bfseries
  \begin{tabular}[#1]{@{}l@{}}#2\end{tabular}%
}

\usepackage{hyperref}
\hypersetup{bookmarksopen,bookmarksnumbered,
pdfpagemode=UseOutlines,
colorlinks=true,
linkcolor=blue,
anchorcolor=blue,
citecolor=blue,
filecolor=blue,
menucolor=blue,
urlcolor=blue
}

\usepackage{geometry}
\begin{document}
%
\title{Benchmarking Robot Manipulation \\
with the Rubik's Cube}
%
%
%

\author{Boling Yang$^{1}$, Patrick E. Lancaster$^{1}$, Siddhartha S. Srinivasa$^{1}$, and Joshua R. Smith$^{1,2}$%
\thanks{Manuscript received: August, 15, 2019; Revised: November, 18, 2019; Accepted: December, 11, 2019.}
\thanks{This paper was recommended for publication by Editor Han Ding upon evaluation of the Associate Editor and Reviewers' comments.}
\thanks{This work was (partially) funded by the National Institute of Health R01 (\#R01EB019335), National Science Foundation CPS (\#1544797), National Science Foundation NRI (\#1637748), National Science Foundation EFMA (\#1832795), National Science Foundation CNS (\#1823148), the Milton and Delia Zeutschel Professorship, the Office of Naval Research, the RCTA, Amazon, and Honda. $^{1}$Boling Yang, Patrick E. Lancaster, Siddhartha S. Srinivasa, and Joshua R. Smith are with The Paul G. Allen School of Computer Science and Engineering, University of Washington, Seattle, WA, USA
        {\tt\footnotesize \{bolingy, planc509, siddh, jrs\}@cs.uw.edu}}%
\thanks{$^{2} $Joshua R. Smith is also with Electrical and Computer Engineering Department, University of Washington, Seattle, WA, USA}%
\thanks{Digital Object Identifier (DOI): see top of this page.}
}
%
%

\markboth{IEEE Robotics and Automation Letters. Preprint Version. Accepted January, 2020}
{Yang \MakeLowercase{\textit{et al.}}: Benchmarking Robot Manipulation with the Rubik's Cube} 

%



\maketitle

\begin{abstract}
Benchmarks for robot manipulation are crucial to measuring progress in the field, yet there are few benchmarks that demonstrate critical manipulation skills, possess standardized metrics, and can be attempted by a wide array of robot platforms. To address a lack of such benchmarks, we propose Rubik's cube manipulation as a benchmark to measure simultaneous performance of precise manipulation and sequential manipulation. The sub-structure of the Rubik's cube demands precise positioning of the robot's end effectors, while its highly reconfigurable nature enables tasks that require the robot to manage pose uncertainty throughout long sequences of actions. We present a protocol for quantitatively measuring both the accuracy and speed of Rubik's cube manipulation. This protocol can be attempted by any general-purpose manipulator, and only requires a standard 3x3 Rubik's cube and a flat surface upon which the Rubik's cube initially rests (e.g. a table). We demonstrate this protocol for two distinct baseline approaches on a PR2 robot. The first baseline provides a fundamental approach for pose-based Rubik's cube manipulation. The second baseline demonstrates the benchmark's ability to quantify improved performance by the system, particularly that resulting from the integration of pre-touch sensing. To demonstrate the benchmark's applicability to other robot platforms and algorithmic approaches, we present the functional blocks required to enable the HERB robot to manipulate the Rubik's cube via push-grasping.
\end{abstract}


%
\IEEEpeerreviewmaketitle

\section{Introduction}
\label{sec:intro}

Despite agreement among researchers that quantitative evaluations are vital for measuring progress in the field of robot manipulation, widely-adopted benchmarks have remained elusive. Related fields such as object recognition, object tracking, and natural language processing use standardized datasets as proxies for quantifying how algorithms will perform in the real-world. Developing benchmarks that serve as sufficiently representative proxies is a challenge for all of these fields. This is even more challenging for the field of robot manipulation because it is no longer sufficient to just observe the environment - instead the robot must interact with the environment to achieve a task. This additional obstacle may explain why there are only a limited number of benchmarks for robot manipulation that possess clear, quantifiable measures, demonstrate aspects of real-world manipulation, and can be attempted by a wide array of robot platforms.

\begin{figure}[!t]
\centering
\includegraphics[width=0.49\textwidth]{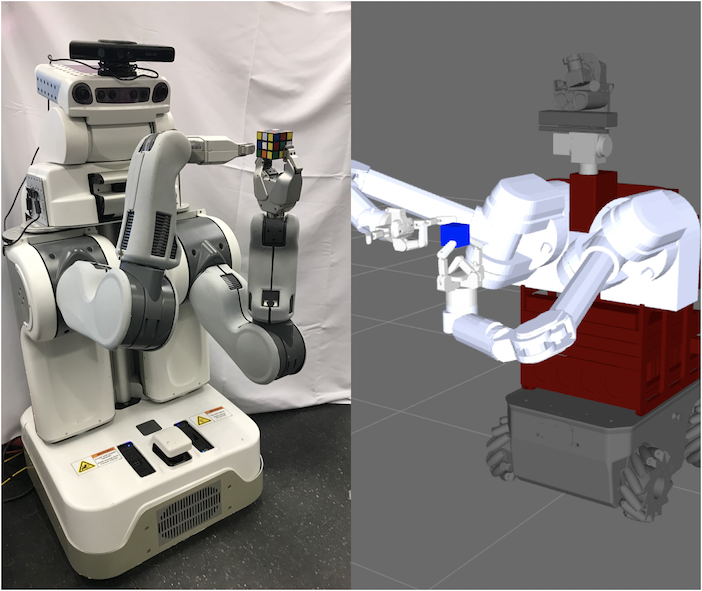}
\caption{Rubik's cube manipulation can be used to benchmark robot manipulation across a wide array of algorithmic approaches and robot platforms, such as the PR2 and HERB.}
\label{fig:best}
\end{figure}

The YCB Object and Model Set~\cite{calli2015benchmarking} establishes standards for both specifying and reporting the results of benchmarks, and facilitates the distribution of a wide variety of manipulation objects and their corresponding models. It also proposes six robot manipulation benchmarks. A subset of these benchmarks (e.g. Box and Blocks, Block Pick and Place) require the robot to \textit{precisely} position its end-effector with respect to the manipulation object. A distinct subset of benchmarks (e.g. Pitcher-Mug, Table Setting) require the robot to perform short \textit{sequences} of manipulation in which later manipulations are heavily influenced by earlier ones. While all of the benchmarks demonstrate critical aspects of robot manipulation, none of them strenuously measure the robot's ability to \textit{simultaneously} perform precise manipulation and sequential manipulation. We therefore propose a Rubik's Cube manipulation benchmark that requires the robot to perform long sequences of highly precise manipulations, as illustrated by Fig. \ref{fig:best}.

\begin{figure*}[!t]
\centering
\includegraphics[width=0.82\linewidth]{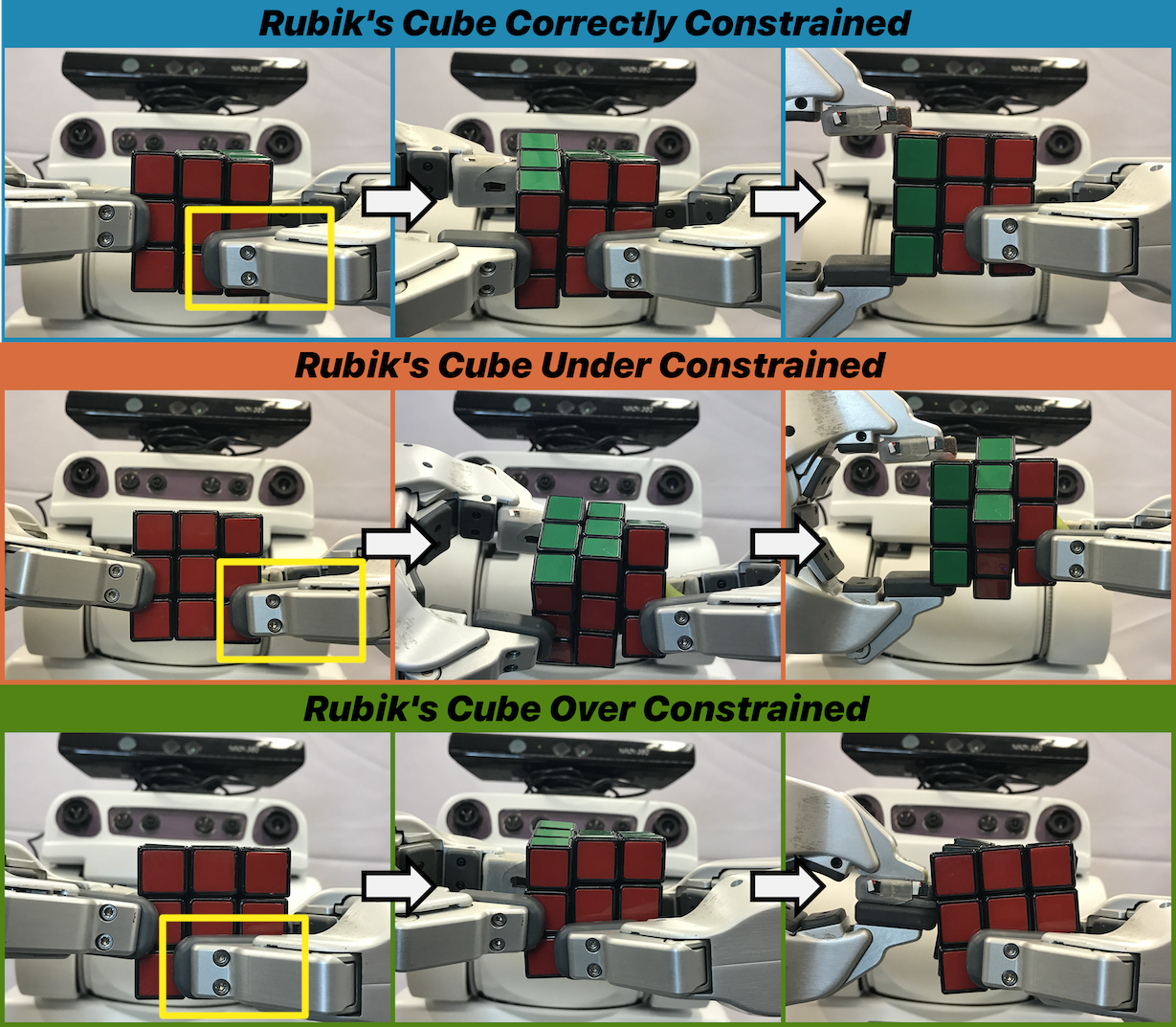}
\caption{The robot must precisely position its grippers to rotate the left column of the Rubik's cube while constraining the middle and right columns in place.
\textbf{Top Row:} The robot correctly positions its grippers: it is constraining the two right columns of the cube.  The yellow box highlights the position of the constraining gripper.
\textbf{Middle Row:} The robot only touches one column of the Rubik's cube and therefore fails to constrain its middle column.
\textbf{Bottom Row:} The right gripper is touching all three columns of the cube; this prevents the left gripper from rotating the left column of the cube.
}
\label{fig:summary}
\end{figure*}

The precision required to manipulate the Rubik's cube is a product of its own structure. Each of its six faces consists of nine sub-cubes arranged into three rows and three columns. The state of the cube can be altered by rotating one of the rows (columns) around an axis parallel to the columns (rows). However, given a row (column) to rotate, both of the other rows (columns) of the face must be held in place. Since each sub-cube has a dimension of only 1.9 cm , each rotation requires the robot to position its end-effectors with sub-centimeter accuracy. Achieving this level of manipulation accuracy is difficult in the presence of uncertainty in the Rubik's cube's pose with respect to the end-effectors. Uncertainty will exist to varying degrees for all general-purpose robots, and can stem from the inability to perfectly calibrate high degree of freedom arms, imperfect actuators, and a number of other sources. As shown in Fig. \ref{fig:summary}, insufficiently \textit{precise} manipulation can cause \textit{under-constrained} portions of the cube to be unintentionally rotated, or prevent intended rotations of \textit{over-constrained} portions of the cube from being executed.

Moreover, Rubik's cube manipulation is more than just a single, precise rotation; for example, solving the Rubik's cube can require up to twenty rotations \cite{rokicki2014diameter}. For each manipulation, there will be some degree of mismatch between the robot's intended and actual execution, resulting in error in the robot's estimate of the Rubik's cube's pose. If the robot does not use sensor feedback, take actions to constrain the pose of the cube, or employ some other method to suppress or respond to errors, then these errors will accumulate and eventually cause manipulation failure. 

The merits of using Rubik's cube manipulation as a benchmark extend beyond being demonstrative of precise and sequential manipulation. In particular, the challenges presented by Rubik's cube manipulation have the potential to be addressed by advances in a wide array of sub-fields, including planning, perception, and control. Although the benchmark directly measures the performance of the system overall, improvements in manipulability resulting from new algorithms and modules can be quantified by comparing to baseline systems. Apart from the research community, the task of Rubik's cube solving is interpretable, familiar, and even entertaining to the general public. With respect to practicality, our proposed benchmark requires minimal setup. Besides the robot itself, the only required items are a standard 3x3 Rubik's cube, and a surface to initially rest the Rubik's cube upon. This simple setup allows the benchmark to be attempted by a wide variety of robot platforms.

We aim to develop a Rubik's Cube manipulation benchmark that objectively measures performance, demonstrates critical aspects of robot manipulation, and can easily be attempted by all types of general-purpose robot manipulators. With these goals in mind, we make the following contributions:

\begin{itemize}
    \item A protocol for measuring the manipulation \textit{accuracy} and \textit{speed} of the Rubik's cube
    \item Two baseline approaches for attempting the benchmark
    \item Open source software for standardization and validation of the benchmark
\end{itemize}

\section{Related Work}

Although a number of researchers have examined benchmarking for robot manipulation \cite{wisspeintner2009robocup,yokokohji2003toy,ulbrich2011opengrasp,behnke2006robot}, exploration of using Rubik's cube manipulation for benchmarking has been extremely limited. Instead, Rubik's cube manipulation has more commonly been used by individual studies to demonstrate the capabilities of new robot algorithms and hardware. In this section, we will first review these studies, and then consider the extent to which Rubik's cube manipulation has previously been proposed as a benchmark for robot manipulation.

Zieliski et. al. \cite{zieliski2007mrroc++} use Rubik's cube manipulation to evaluate their proposed controller architecture for dual arm service robots. They thoroughly describe their methods for identifying visual features to localize the Rubik's cube, thresholding in HSV space to identify colors, and visual-servoing to grasp the Rubik's cube. They also use somewhat specialized hardware; trough-shaped fingers allow the robot's gripper to conform to the corners of the Rubik's cube. Although they report relevant metrics about the individual components of their system, a quantitative evaluation of the overall system's Rubik's cube manipulation performance is not given. 

OpenAI et al.\cite{akkaya2019solving} evaluates the in-hand dexterity of their robotic manipulation system by solving a Rubik's cube. The robot learns in-hand manipulation skills for a five-fingered humanoid hand via reinforcement learning and automatic domain randomization. A customized Rubik's cube embedded with joint encoders allows the robot to estimate the Rubik's cube's state, and  an array of cameras facilitates tracking of the Rubik's cube's position. The in-hand manipulation system achieves a 20\% success rate when solving well scrambled Rubik's cubes.

Higo et. al. \cite{higo2018rubik} present a system for manipulating the Rubik's cube using a dexterous, multi-fingered hand and high-speed vision. The Rubik's cube rests on a flat surface in front of the disembodied gripper throughout the manipulation. Manipulation is decomposed into three motion primitives: yaw rotation, pitch rotation, and rotation of the top row of the Rubik's cube. They report that performing a sequence containing a single instance of each of these motion primitives requires less than one second. With respect to manipulation accuracy, the system succeeded in seven out of ten trials, where each trial consisted of thirty seconds of continuous manipulation.  

Through manipulation of the Rubik's cube, Yang et. al. \cite{yang2017pre} analyze the ability of optical time-of-flight pre-touch sensors to reduce object pose uncertainty during sequential manipulation. Specifically, they compare the error in the robot's estimate of the Rubik's cube's pose when using pre-touch sensors to that of a system that does not use sensor feedback. Both of these approaches serve as foundations for two of the baselines reported in this work, and will be detailed further in Section \ref{sec:baselines}. Yang et. al. note that pose error greater than half the dimension of a sub-cube can cause the robot to under-constrain or over-constrain the Rubik's cube. They find that the pre-touch based approach consistently keeps the pose error below this threshold, while the sensorless approach often exceeds it. Additional experiments and Lancaster et. al. \cite{lancaster2017improved} demonstrate that the pre-touch based approach used to manipulate the Rubik's cube can be extended to general object geometries. 

Each of these studies uses Rubik's cube manipulation as a mechanism for evaluating their technical contributions, rather than establishing a standard benchmark for other researchers to attempt. However, the foundations for Rubik's cube manipulation as a community-wide benchmark do exist. In particular, the YCB Object and Model Set \cite{calli2015benchmarking} contains a Rubik's cube, but it does not specify any related protocols or baselines. Similarly, Zieliski \cite{zielinski2006rubik} argue the merits of using Rubik's cube manipulation as a benchmark and describe requirements of systems that might attempt such a benchmark, but does not present a standard process for evaluating Rubik's cube manipulation. The remaining sections of this article will focus on developing a protocol generalizable to a wide variety of robot platforms, and establishing baseline scores to which other researchers can compare their systems. 
\section{Protocol for Rubik's Cube Manipulation}
We propose a protocol to measure both the accuracy and speed of Rubik's cube manipulation. The robot is required to perform a sequence of Rubik's cube manipulations as quickly as possible. The accuracy of the manipulation is quantified by the number of successful manipulations that can be achieved, while the manipulation speed is inversely proportional to the amount of time required to execute the sequence. We organize our benchmark into multiple tiers. Each tier is denoted as Rubiks-M-N. Rubiks-M-N consists of M \textit{consecutive} trials, where in each trial the robot must pick the Rubik's cube up off of the table and complete N rotations. Each of the M trials prescribes a distinct manipulation sequence of length N.

\subsection{Protocol Prerequisites}
\label{sec:protocol-prereqs}
The protocol requires a standard 3x3 Rubik's cube of dimension 5.7 centimeters along each edge and a flat platform on which to place the cube. We recommend using the inexpensive and widely available 3x3 Hasboro Gaming Rubik's Cube, item number A9312. Researchers and/or the robot can choose to use the flat platform to decrease uncertainty or extend manipulability. The Rubik's cube should initially be positioned to rest on top of the table's surface such that its center is located at the middle of the robot's workspace in the x and y directions (with respect to the robot's base frame). It should be oriented such that its top face is parallel to the ground, and its back face perpendicular to the sagittal plane of the robot. The robot's manipulator(s) should not initially make contact with the cube. Once the robot has started the benchmarking process, no form of human intervention is allowed.

We provide software to ensure benchmark consistency across research groups, and to aid in validation of the achieved score. The first module provides pseudo-random sequences of Rubik's cube rotations with a fixed seed. The generated sequences are specified with  \href{https://ruwix.com/the-rubiks-cube/algorithm}{standard Rubik’s cube notation}. The second module outputs the final expected state of the Rubik's cube after being given its initial state and a sequence of rotations to perform. The source code and instructions for usage can be found here: \href{https://gitlab.cs.washington.edu/bolingy/rubiks-cube-benchmark}{https://gitlab.cs.washington.edu/bolingy/rubiks-cube-benchmark}

\subsection{Protocol Details}
This protocol measures the robot's ability to perform a sequence of Rubik's cube manipulations as quickly as possible. We propose twelve specific tiers for the experimenter to attempt: Rubiks-1-5, Rubiks-1-10, Rubiks-1-20, Rubiks-1-50, Rubiks-1-100, Rubiks-1-200, Rubiks-5-5, Rubiks-5-10, Rubiks-5-20, Rubiks-5-50, Rubiks-5-100, and Rubiks-5-200.  For example, in Rubiks-5-100, the robot is required to pick up the Rubik's cube and perform a 100-rotation sequence of Rubik's cube manipulations five times in a row. The first six tiers provide an optimistic evaluation of the system's capability. The later six tiers reflect the robustness of the system's performance. Note that completing Rubiks-1-20 signifies that the robot has sufficient manipulation accuracy to solve any Rubik's cube.

The protocol consists of the following steps:
\begin{enumerate}
    \item Experimenter or the robot decides which tier to attempt
    \item Robot acquires manipulation sequence from the provided software
    \item Experimenter places the Rubik's cube on the surface in front of the robot as defined in Sub-section \ref{sec:protocol-prereqs}
    \item Robot picks up cube and begins to execute the manipulation sequence
    \item Robot terminates manipulation
    \item Experimenter validates that the final cube state is correct using the provided software
    \item Return to step two if there are remaining trials to be completed
\end{enumerate}

For each trial, if the final cube state is correct, the system's score is the time elapsed between the robot first making contact with the Rubik's cube and the termination of manipulation. The final score for the tier is the average trial score and standard deviation. The experimenter should report any completed tiers, corresponding speed scores, and clear video recording of these scores being attained. For a given value of M, completing higher tiers (larger N) indicates higher manipulation accuracy, while completing a given tier faster demonstrates better manipulation speed. For a given value of N, completing a tier with larger M indicates better robustness.

\section{Baselines for Rubik's Cube Manipulation}
\label{sec:baselines}

In order to provide initial metrics for comparison, we attempt and report the results of the benchmark for two distinct approaches. Note that each of the reported baselines uses a bi-manual approach, requiring the robot to perform long sequences of re-grasps in the presence of object pose uncertainty. In the first baseline, a PR2 robot initially localizes the Rubik's cube with its head-mounted camera, but then attempts to perform sequences of manipulation without any further sensor feedback. While the initial cube pose is also obtained through the head-mounted camera, the second baseline outfits the PR2 robot's end-effectors with optical pre-touch sensors, allowing it to re-localize the Rubik's cube just prior to each re-grasp. The following sub-sections will describe each baseline in greater detail, and then report the benchmark scores achieved with each baseline. Finally, we present empirical evidence that our benchmark can be attempted by other robot platforms, particularly (but not limited to) the HERB \cite{Srinivasa2010herb} robot.

\subsection{Dead Reckoning Baseline}
\label{sec:dead_reckoning}

This baseline \cite{yang2017pre} serves as a foundation for synthesizing a sequence of Rubik's cube rotations into corresponding trajectories that can be executed on a robot. The PR2 robot initially estimates the pose of the Rubik's cube with a head-mounted camera, but then assumes that it is able to position its grippers with perfect accuracy. The given sequence of rotations guides the robot through a finite state machine\footnote{This state machine maps a sequence of Rubik's cube rotations to a sequence of robot poses; the source code can be found here: https://gitlab.cs.washington.edu/bolingy/rubiks-cube-state-machine}. For each rotation, this finite state machine encodes the pose trajectories that the robot should execute in order to achieve the rotation given the current configuration of its end effectors. The baseline's main weakness is that it assumes perfect knowledge of the pose of the Rubik's cube, and its results (Section \ref{sec:results}) highlight the need to examine additional baselines that consider pose uncertainty during long sequences of manipulation.

\subsection{Fingertip Sensor Aided Baseline}
To reduce the pose uncertainty observed in Section \ref{sec:dead_reckoning}, the second baseline continuously re-estimates the pose of the Rubik's cube. Before making contact, the robot will use optical time-of-flight proximity sensors mounted on its grippers to re-estimate the Rubik's cube's pose. The finite state machine in Section \ref{sec:dead_reckoning} uses these pose estimates to appropriately adjust the robot's pre-grasps. Yang et. al. \cite{yang2017pre} demonstrate that this sensing method reduces the robot's end effector positioning error to less than half of a centimeter.

\subsection{Baseline Results}
\label{sec:results}
The dead reckoning baseline completed Rubiks-1-20 in 463.45 seconds, and it was unable to complete any of the higher single trial tiers. This result indicates that the system has sufficient accuracy to solve a Rubik's cube given multiple attempts. On the other hand, the dead reckoning baseline was only capable of completing the lowest tier for five consecutive trials, with an average manipulation time of 113.35 seconds and standard deviation of 17.16 seconds. Without constant object state estimation, the system is not robust against the uncertainty accumulated throughout sequential manipulation. The robot's estimation of the object's pose is largely inaccurate after 5 rotations of the Rubik's cube, preventing it from completing 10 or more rotations across 5 consecutive trials.

Relative to the dead reckoning baseline, the fingertip sensor aided baseline achieved better performance with respect to both speed and accuracy in single trial Rubik's Cube manipulation. It successfully finished Rubiks-1-20 in 447.96 seconds, and was able to complete Rubiks-1-100 in 2207.97 seconds. This system also demonstrated much better robustness by successfully completing Rubiks-5-20. However, when comparing the Rubiks-5-5 results between both baselines, the addition of sensing actions in the latter baseline results in larger variance in speed protocol score. All of the tiers completed by our baselines and corresponding scores are reported in Table \ref{tbl:baselines_scores}.

\begin{table}[!t]
\centering
\begin{tabular}{l|ll}
\toprule
\diagbox[width=10em]{\textbf{Tier}}{\textbf{Baseline}} &
\specialcellbold{Dead Reckoning\\PR2 (Avg/Std Dev)(s)} &
\specialcellbold{Sensor Aided\\PR2 (Avg/Std Dev)(s)} \\
\midrule
\textbf{Rubiks-1-5}   & 139.07 / - & 126.67 / -\\
\textbf{Rubiks-1-10}  & 248.43 / - & 215.37 / -\\
\textbf{Rubiks-1-20}  & 463.45 / - & 447.96 / -\\
\textbf{Rubiks-1-50}  &  -- & 1123.04 / -\\
\textbf{Rubiks-1-100} & -- & 2207.97 / -\\
\textbf{Rubiks-1-200} & -- & --  \\
\midrule
\textbf{Rubiks-5-5}   & 113.35 / 17.16 & 113.81 / 18.35\\
\textbf{Rubiks-5-10}  & --  & 214.71 / 30.39\\
\textbf{Rubiks-5-20}  & --  & 416.61 / 23.10\\
\textbf{Rubiks-5-50}  &  -- & -- \\
\textbf{Rubiks-5-100} & -- & --  \\
\textbf{Rubiks-5-200} & -- & --  \\
\bottomrule
\end{tabular}
\caption{The scores (mean and standard deviation of manipulation time) achieved by our baseline approaches on each of the tiers. Each row corresponds to a separate tier, Rubiks-M-N. Rubiks-M-N  consists of M consecutive trials, where in each trial the robot must pick the Rubik’s cube up off of the table and complete N rotations.}
\label{tbl:baselines_scores}
\end{table}

\subsection{Extending Beyond Baselines}
\begin{figure}[!t]
\centering
\includegraphics[width=0.49\textwidth]{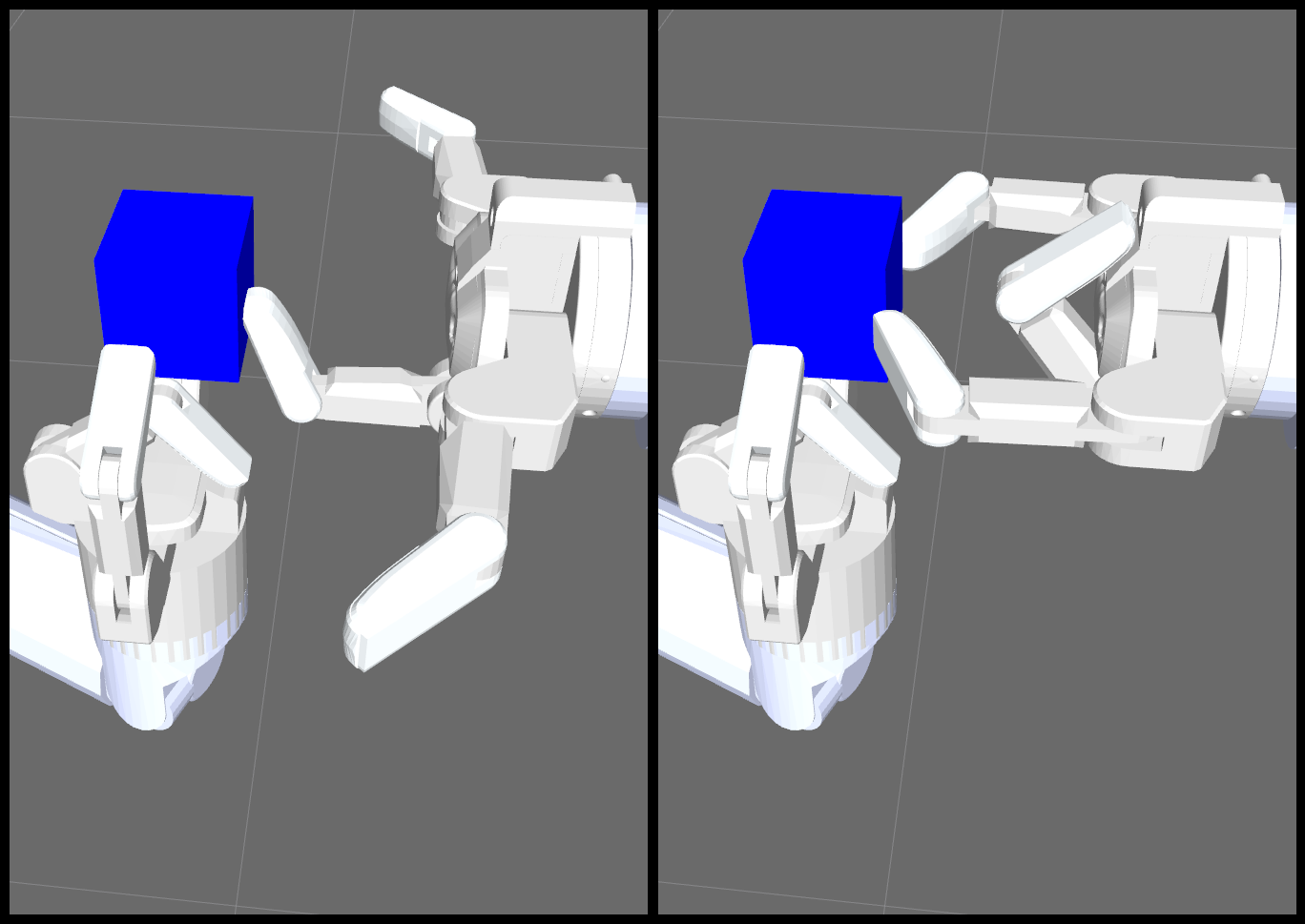}
\caption{HERB's simulated Barrett hands manipulate the Rubik's cube in blue. \textbf{Left:} The right gripper uses a push-grasp to make contact with the Rubik's cube. Upon making contact, the right gripper reaches the desired pre-grasp. \textbf{Right:} The right gripper closes its outer fingers to transition from pre-grasp to grasp.}
\label{fig:HERBgrasp}
\end{figure}

In this section, we demonstrate the feasibility of applying our benchmark to other manipulation paradigms and robot platforms. In particular, we explore the use of push-grasping on the HERB robot. Push grasping is another method that allows the robot to reduce pose uncertainty during manipulation \cite{dogar2011framework}. Here, we demonstrate in simulation that HERB is able to execute the push-grasps necessary to solve the Rubik's cube. The finite state machine used in Section \ref{sec:dead_reckoning} is modified such that each re-grasp of the Rubik's cube is achieved through a push grasp. Specifically, given a Rubik's cube face to grasp, the HERB robot's gripper approaches perpendicularly to the face. The robot's middle finger is positioned such that once it has made contact with the face, the robot's gripper has reached a pre-grasp amenable to rotating the face, as shown in Fig \ref{fig:HERBgrasp}. The simulated robot is indeed capable of performing all of the trajectories prescribed by the state machine, demonstrating the generalizability of our benchmark beyond the provided baselines.

\section{Discussion}
We have developed a benchmark that can measure precise, sequential manipulation across a wide variety of robot platforms. However, others may disagree with our design choices. In particular, the benchmark measures the overall accuracy and speed at which the Rubik's cube is manipulated, but there are no statistics that quantify the performance of the individual manipulation actions in the sequence. Such statistics were omitted for practicality. For instance, it would be difficult for the experimenter to make manual measurements without interfering with the manipulation process. Furthermore, external systems for collecting ground-truth (e.g. motion capture) can become occluded, and are difficult to standardize across research groups. 

There are a number of research directions beyond the reported baselines to which our benchmark can be applied. In particular, basic visual servoing methods may experience a large amount of occlusion during manipulation. Our benchmark could be used to evaluate the effectiveness of visual servoing methods that specifically address the challenges presented by occlusion during manipulation \cite{malis2005unified,lippiello2007position,shi2018adaptive }. 

Along a different direction, methods for planning under uncertainty can act less conservatively than our fingertip sensor aided baseline. Instead of prescribing uncertainty reduction measures prior to each re-grasp, uncertainty-aware planners will be able to more judiciously determine when uncertainty reduction is necessary \cite{ong2010planning, bry2011rapidly, van2012motion}. The improvement in system performance induced by such algorithms could be measured using our benchmark. Finally, as suggested by Ma and Okamura \cite{ma2011dexterity, okamura2000overview}, robots with versatile kinematics and/or highly dynamic capabilities are key components of truly dexterous manipulation. In particular, high degrees of in-hand dexterity have been achieved through the use of external force \cite{dafle2014extrinsic}, optimal control \cite{kumar2016optimal}, and deep reinforcement learning \cite{andrychowicz2018learning}. Our benchmark can serve as a metric to evaluate the robustness and agility of these methods relative to one another.

\ifCLASSOPTIONcaptionsoff
  \newpage
\fi



%
\newpage
\bibliography{references}
\bibliographystyle{IEEEtran}

\end{document}